\documentclass[letterpaper]{article}
\usepackage{aaai2026}
\nocopyright  

\usepackage{times}
\usepackage{helvet}
\usepackage{courier}
\usepackage[hyphens]{url}
\usepackage{graphicx}
\usepackage{natbib}
\usepackage{caption}
\usepackage{booktabs}   
\usepackage{multirow}   
\usepackage{amsmath}    
\usepackage{amssymb}    
\usepackage{xcolor}     
\usepackage{colortbl}   
\usepackage{subcaption} 

\title{The Visual Insensitivity Gap: \\ Diagnosing When Vision-Language Models Fail to Use Visual Evidence}

\author{
    Genpei Zhang
}
\affiliations{
    University of Wisconsin--Madison\\
    genpei.zhang@wisc.edu
}

\begin{document}

\maketitle

\begin{abstract}
Vision-language models are evaluated by aggregate accuracy on multimodal benchmarks, a practice that implicitly assumes the model uses its visual input. We show this assumption fails on $40$\%--$97$\% of samples across six VLMs and three perceptual benchmarks: blurring the question-relevant visual region leaves the next-token distribution nearly unchanged. We name this phenomenon the \emph{Visual Insensitivity Gap} and quantify it with a per-sample Visual Sensitivity Index (VSI). The gap is a property of samples, not of models: VSI ranks correlate across models (grand-mean Spearman $\rho{=}{+}0.40$, permutation $p<10^{-3}$), so the same samples are flagged insensitive by VLMs sharing no architectural detail beyond a contrastively pretrained vision tower. The mechanism is concrete: on the insensitive samples, a linear probe on each model's \emph{own} vision tower distinguishes perturbed from clean images at $0.72$--$0.79$ accuracy, yet the model's argmax token changes on only $2$\%--$11$\% of the same samples, an encoder--LLM gap above $0.65$ on every model. Mapping VSI's diagnostic utility cell by cell surfaces a strong regime (multi-choice reasoning on capable VLMs: $\mathrm{AUROC}{=}0.85$--$0.87$) and a weak regime (well-calibrated factuality, where softmax confidence already leads). VSI is not a universal best abstention signal; it is a sample-intrinsic indicator of vision-ignoring failure, best used as a conditional ensemble component.
\end{abstract}

\section{Introduction}\label{sec:intro}

Vision-language models (VLMs) are routinely evaluated by their aggregate accuracy on multimodal benchmarks such as POPE~\citep{li2023pope}, MMVP~\citep{tong2024eyes}, HallusionBench~\citep{guan2024hallusion}, and MMStar~\citep{chen2024mmstar}. This practice that implicitly assumes the model uses its visual input. Yet across six recent VLMs spanning three architecture families~\citep{liu2024llava15,liu2024llavanext,bai2025qwen3vl,bai2025qwen25vl,laurencon2024idefics3}, perturbing the question-relevant visual region leaves the output unchanged on $40$\%--$97$\% of perceptual-benchmark samples, even when a linear probe~\citep{alain2017probing} on the model's own vision encoder detects the perturbation. The gap matters wherever a VLM's answer carries decision weight (clinical triage, document-grounded question answering, embodied agents): a confidently wrong answer that ignored the visual evidence is a different failure from a hedged answer that engaged with it, yet aggregate accuracy cannot tell them apart.

We name this phenomenon the \emph{Visual Insensitivity Gap}: the discrepancy between what the vision encoder represents and what the language head writes out. We quantify it with the Visual Sensitivity Index (VSI), a per-sample measure of how much the next-token distribution moves when the question-relevant visual region is blurred. VSI distributions show a consistent heavy left tail across the six models and four benchmarks we study: a sub-population of insensitive samples separates from a bulk of sensitive ones. The sub-population is sample-intrinsic: VSI ranks agree across models with a grand-mean Spearman rank correlation of $\rho{=}+0.40$ on perceptual benchmarks (permutation $p<10^{-3}$ on every tested pair), so the same samples tend to be flagged insensitive by VLMs that share no architectural detail beyond a contrastively pretrained~\citep{radford2021learning} vision tower. Visual insensitivity is therefore a property of samples and of the dominant pretraining paradigm, not of any individual model.

Mechanism follows. On samples a model flags as insensitive, a linear probe on that model's own vision tower distinguishes perturbed from unperturbed images at $0.72$--$0.79$ accuracy (a frozen reference CLIP probe reads $0.82$). The model's argmax token, by contrast, changes on only $2$\%--$11$\% of these same samples. The encoder--decoder gap of $0.66$--$0.71$ replicates on all six models. The vision tower sees the perturbation; the language head, on a large fraction of samples, does not propagate that signal. This measurement rests on an input-level interventional framing, which we adopt throughout: attention-based interpretability of VLMs has produced rich descriptive maps~\citep{neo2024towards,stan2024lvlm} but does not by itself establish whether the represented information is used in producing the output. Removing visual evidence and watching the softmax move makes the causal question concrete: would the answer change if this evidence were absent? The Visual Insensitivity Gap is what we see when, sample after sample, the answer is no.

The gap is also diagnostic. Samples with low VSI but high softmax confidence form a quadrant of \emph{confidently wrong while ignoring vision} predictions; these track hallucination by every measure we examine. On multi-choice reasoning (MMStar), the math and science subcategories on Qwen2.5-VL-32B yield $\mathrm{AUROC}_{\mathrm{VSI}}{=}0.85$ and $0.87$ respectively, the strongest single-signal selective-generation results we observe. Yet VSI is not a universal best signal: on POPE-style factuality, softmax confidence still leads on several model--benchmark cells. Our framing is therefore conditional: where VSI helps is itself an empirical, per-cell question.

Our thesis is that visual insensitivity is a robust, sample-intrinsic property of contemporary VLMs, and we test it in three complementary ways. We (1) formalise the Visual Insensitivity Gap and establish its heavy-tailed, cross-model structure on a six-model, four-benchmark grid (\S\ref{sec:vig}); (2) trace the mechanism to an encoder--LLM disconnect that replicates on all six models (\S\ref{ssec:disconnect}); and (3) map where the per-sample signal is diagnostically useful (strong on multi-choice reasoning, weak on well-calibrated factuality), and how it combines with softmax and verbalised-confidence baselines (\S\ref{sec:diagnostic}--\S\ref{sec:cond}). We close by discussing what these findings do, and do not, tell us about how VLMs use vision.
\section{The Visual Insensitivity Gap}\label{sec:vig}

This section formalises the phenomenon and traces it to a measurable encoder--LLM disconnect.

\subsection{The Visual Sensitivity Index}\label{ssec:vsi-def}

For a vision-language model $f$ producing next-token distributions $f(\cdot \mid x,q)$ given image $x$ and question $q$, we measure visual sensitivity by perturbing the image and observing how the output distribution moves. Let $x_\sigma$ denote the image with a Gaussian-blur perturbation of strength $\sigma$ applied to the region most relevant to the question $q$. The Visual Sensitivity Index is
\[
\mathrm{VSI}(x,q;f) \;=\; \mathrm{KL}\!\left(f(\cdot \mid x, q) \;\|\; f(\cdot \mid x_\sigma, q)\right),
\]
with $\sigma{=}20$ as the default (robustness across $\sigma\in\{10,20,40\}$ in \S\ref{ssec:robustness}). The KL form captures the full distributional shift, not only argmax flips; localising the perturbation isolates whether the model uses the \emph{specific} content the question asks about, not images in general.

The question-relevant region is located by parsing the principal noun phrase from $q$, querying Grounding-DINO~\citep{liu2023groundingdino} for an open-vocabulary box, and refining the box to a pixel mask with SAM~\citep{kirillov2023segment}. Samples without a confident box fall back to whole-image perturbation. \S\ref{ssec:robustness} also studies a whole-image variant (\emph{whole-image VSI}). The raw sample pool per benchmark is identical across models; accuracy-dependent metrics use the answer-parseable subset (App.~\ref{app:parsing}), with per-cell counts of $93$ to $500$, and all metrics are computed per (model, benchmark) cell. We report VSI on six VLMs (LLaVA-1.5-7B~\citep{liu2024llava15}, LLaVA-NeXT-Mistral-7B~\citep{liu2024llavanext}, Idefics3-8B~\citep{laurencon2024idefics3}, Qwen3-VL-8B~\citep{bai2025qwen3vl}, Qwen2.5-VL-7B and Qwen2.5-VL-32B~\citep{bai2025qwen25vl}) on three perceptual benchmarks (POPE~\citep{li2023pope}, MMVP~\citep{tong2024eyes}, HallusionBench~\citep{guan2024hallusion}) and one multi-choice reasoning benchmark (MMStar~\citep{chen2024mmstar}).

\subsection{A heavy left tail of insensitive samples}\label{ssec:bimodality}

\begin{figure*}[t]
  \centering
  \includegraphics[width=\linewidth]{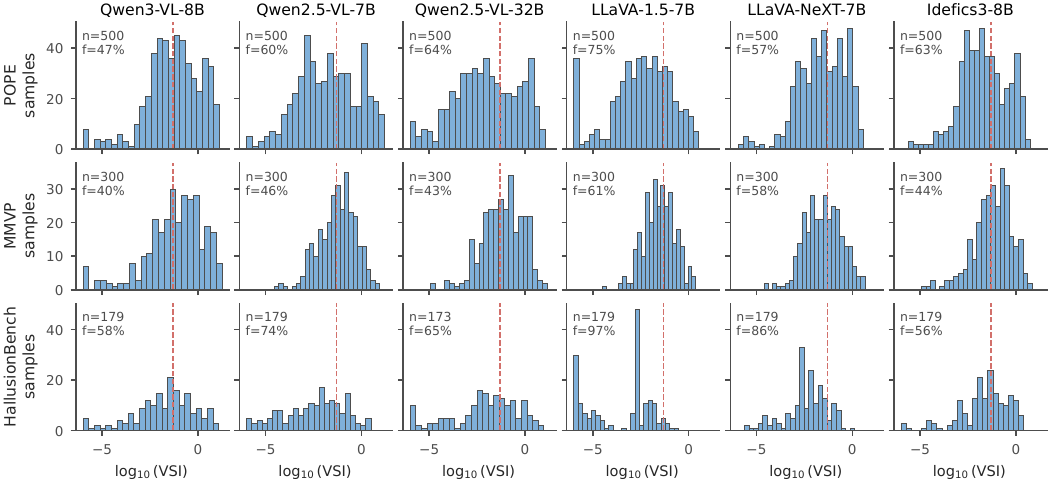}
  \caption{Distribution of per-sample VSI across $6$ models $\times$ $3$ benchmarks. A heavy left tail of insensitive samples (VSI${<}0.05$, red dashed line) separates from a sensitive bulk; the insensitive fraction (\emph{f}) ranges from $40$\% to $97$\% across cells.}
  \label{fig:bimodality}
\end{figure*}

Figure~\ref{fig:bimodality} plots the per-sample VSI histogram for every (model, benchmark) cell. The shape is consistent: a heavy left tail of insensitive samples and a separated bulk of sensitive samples. The insensitive fraction, i.e.\ samples with $\mathrm{VSI}{<}0.05$, spans $40$\% to $97$\%, with HallusionBench typically highest (figures and diagrams reduce the informativeness of local-region perturbation) and POPE typically lowest. The heavy left tail is not an artefact of any one model: every model shows it on every benchmark we tested. The fractions are not monotone in model scale either: Qwen2.5-VL-32B's tail is no lighter than Qwen2.5-VL-7B's. Hartigan's dip test does not reject unimodality and silhouette scores are moderate (Table~\ref{tab:bimodality-stats}), so we describe the structure as a heavy left tail rather than bimodality; the distinction matters because a heavy tail offers no natural threshold, which motivates the quintile-based analyses used throughout.

\begin{table*}[t]
\begin{minipage}[t]{0.48\textwidth}
\centering\footnotesize
\begin{tabular}{lrrrrr}
\toprule
 & & \multicolumn{2}{c}{Fraction with VSI} & & \\
\cmidrule(lr){3-4}
Variant & Median & ${<}0.01$ & ${<}0.05$ & Dip $p$ & Silh. \\
\midrule
Region-blur     & $0.09$ & $24.5$\% & $41.5$\% & $0.99$ & $0.57$ \\
Patch-occlusion & $0.09$ & $21.0$\% & $42.0$\% & $0.96$ & $0.54$ \\
Whole-image     & $2.78$ & $\phantom{0}1.5$\% & $\phantom{0}6.0$\% & $0.24$ & $0.61$ \\
\bottomrule
\end{tabular}
\caption{Per-variant VSI statistics ($n{=}200$, Qwen3-VL-8B). Localised perturbations carry a heavy left tail ($\sim 42$\% below $\mathrm{VSI}{=}0.05$); whole-image blurring almost always moves the softmax. The dip test~\citep{hartigan1985dip} does not reject unimodality; silhouette~\citep{rousseeuw1987silhouette} is moderate.}
\label{tab:bimodality-stats}
\end{minipage}\hfill
\begin{minipage}[t]{0.50\textwidth}
\centering\footnotesize
\begin{tabular}{lcccc}
\toprule
 & \multicolumn{2}{c}{Own-tower probe acc} & & \\
\cmidrule(lr){2-3}
Model & low-VSI & high-VSI & $\Delta$argmax & Gap \\
\midrule
Qwen3-VL-8B           & $0.734{\pm}.03$ & $0.893$ & $0.024$ & $0.710$ \\
Qwen2.5-VL-7B         & $0.764{\pm}.07$ & $0.885$ & $0.060$ & $0.704$ \\
Qwen2.5-VL-32B        & $0.741{\pm}.06$ & $0.889$ & $0.084$ & $0.656$ \\
LLaVA-1.5-7B          & $0.722{\pm}.11$ & $0.859$ & $0.060$ & $0.661$ \\
LLaVA-NeXT-7B & $0.789{\pm}.04$ & $0.906$ & $0.108$ & $0.680$ \\
Idefics3-8B           & $0.753{\pm}.04$ & $0.876$ & $0.084$ & $0.668$ \\
\bottomrule
\end{tabular}
\caption{Encoder--LLM disconnect on the shared $n{=}83$ low-VSI pool, probed on each model's \emph{own} vision tower ($\pm$ = $5$-fold std); ``high-VSI'' is the matched sensitive control. $\Delta$argmax: fraction of samples whose top-1 token changes under perturbation. Gap = probe acc $-$ $\Delta$argmax, ${>}0.65$ throughout. Reference-CLIP probe: $0.82{\pm}0.04$ (App.~\ref{app:probe}).}
\label{tab:disconnect}
\end{minipage}
\end{table*}

The whole-image variant (Table~\ref{tab:bimodality-stats}, last row) is the sanity check: its insensitive fraction collapses to $\approx 6$\%, so the localised heavy tail is not a generic ``any-perturbation'' artefact. Region-level VSI captures \emph{which} visual content the model attends to, not whether it attends to images at all.

\subsection{Visual insensitivity is sample-intrinsic}\label{ssec:sample-intrinsic}

We call a per-sample VSI score \emph{sample-intrinsic} if its rank order is preserved across models above chance: given VSI scores $V^A$ and $V^B$ from models $A,B$ on the same pool, the Spearman rank correlation $\rho(V^A,V^B)$ is bounded away from zero and separated from any permutation null. Without this test, the heavy-tail pattern of \S\ref{ssec:bimodality} could be model-idiosyncratic: each model's tail could comprise different samples, with only the aggregate shape in common.

Over the three perceptual benchmarks, the grand-mean Spearman of region-blur VSI across the $15$ model pairs is $\rho{=}{+}0.40$. Same-family pairs tend to correlate more strongly on average (pooled region-blur mean $\rho{=}0.51$ vs $0.37$ for cross-family pairs, with the same direction on every benchmark). Cross-family agreement nonetheless remains substantial: Qwen3-VL vs LLaVA-1.5 reaches $\rho{=}0.45$ on POPE, and even the weakest pair we tested (Qwen2.5-VL-32B vs Idefics3 on MMVP) retains $\rho{=}{+}0.20$, well separated from a permuted null. Cross-family pairs share no component beyond a contrastively pretrained vision tower; their agreement suggests it is the shared pretraining paradigm that shapes which samples get ignored. A practical corollary, used in \S\ref{ssec:disconnect}, is that an \emph{insensitivity flag} derived from one VLM transfers to others at lower but informative fidelity. Appendix Fig.~\ref{fig:sample-intrinsic-supp} renders joint VSI quintiles for three model pairs; the high-error cluster sits in the low-VSI corner for both models.

Bootstrap $95$\% CIs ($B{=}1000$) lie fully above $\rho{=}0$ for every POPE and MMVP pair; permutation tests confirm (\S\ref{ssec:robustness}). Cross-model agreement is benchmark-dependent: POPE yields a mean $\rho{=}0.55$, MMVP $\rho{=}0.34$, and HallusionBench $\rho{=}0.32$, with HallusionBench weakest because its chart-reading items compete with our region-detection pipeline. VSI thus captures a benchmark-conditional sample property, not a universal one.

\subsection{Encoder--LLM disconnect: the mechanism}\label{ssec:disconnect}

\begin{figure*}[t]
  \centering
  \includegraphics[width=\linewidth]{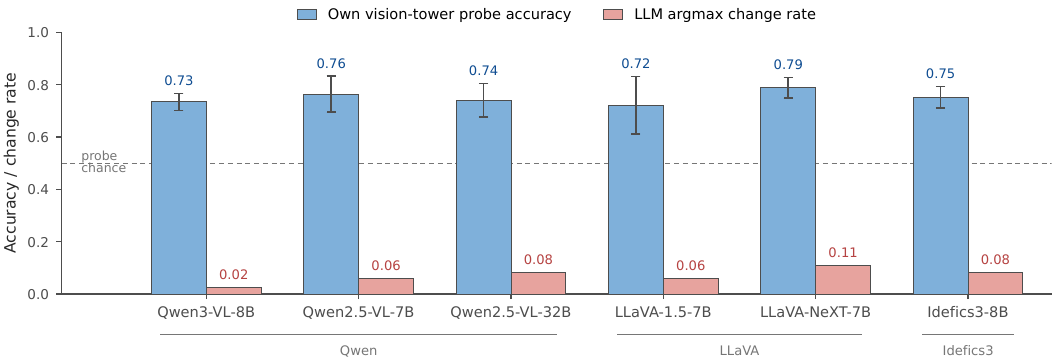}
  \caption{Encoder--LLM disconnect on the low-VSI subset across $6$ VLMs. Blue: linear-probe accuracy on each model's own vision tower ($0.72$--$0.79$; error bars: $5$-fold std). Solid orange: argmax-change rate on the same low-VSI samples ($2$--$11$\%). Light orange: argmax-change rate on the high-VSI samples (where the gap largely closes). The vertical bracket shows the encoder--LLM gap of $0.66$--$0.71$.}
  \label{fig:disconnect}
\end{figure*}

What does a vision encoder represent on insensitive samples? We answer this with a controlled intervention. The six VLMs do not share a vision tower (the LLaVA models use CLIP-ViT-L/14; Idefics3 uses SigLIP-SO400M; the Qwen models use natively trained ViTs), so we probe \emph{each model's own tower}: an $L_2$-regularised logistic-regression linear probe on $\ell_2$-normalised final-layer features, trained to discriminate perturbed from unperturbed images with grouped $5$-fold cross-validation (App.~\ref{app:probe}). On the $83$ samples flagged low-VSI by Qwen3-VL ($\mathrm{VSI}{<}0.05$), a single shared pool evaluated on each model's own tower, every tower linearly separates perturbed from clean images at $0.72$--$0.79$ accuracy (Table~\ref{tab:disconnect}); a frozen reference CLIP encoder, independent of all six models, reads $0.82 \pm 0.04$ on the same pool (App.~\ref{app:probe}).

Yet the same models' outputs barely register the perturbation: the argmax token changes on only $2$\%--$11$\% of the same samples (Table~\ref{tab:disconnect}). The encoder--LLM gap, own-tower probe accuracy minus argmax-change rate, is $0.66$--$0.71$ on every model. The reference-encoder gap ($0.72$--$0.80$) agrees in direction and magnitude, so the disconnect does not depend on whose representation is probed. The high-VSI control shows the asymmetry is not probe weakness: there, probe accuracy rises to $0.86$--$0.91$ and the argmax-change rate rises with it (Fig.~\ref{fig:disconnect}). The disconnect is therefore not a property of the architecture as a whole; it is concentrated where VSI is small.

Three alternative explanations are worth ruling out, all by the same data. \emph{Too-weak perturbation}: ruled out by the $0.72$--$0.79$ own-tower probe accuracy ($0.86$--$0.91$ on the high-VSI control), well above the $2$--$11$\% argmax-change rate. \emph{Greedy decoding hiding distributional shift}: ruled out because VSI is the KL between the two full softmax distributions and is small on the low-VSI subset by construction. \emph{Uniform reliance on a fixed visual prior}: ruled out because the gap is concentrated in the bottom VSI quintile and closes toward the top, so the same language head \emph{does} respond to visual changes on other samples from the same distribution. The disconnect is therefore best characterised as an input--output usage gap, consistent with a routing failure rather than a representational deficit: every model's own vision tower, and a frozen generic encoder, carries the information, and the same language head demonstrably responds to visual change on high-VSI samples; on the affected samples, the path from encoder to output is simply silent.
\section{Diagnostic Value}\label{sec:diagnostic}

A heavy left tail on its own is a phenomenon, not a diagnostic. This section examines whether VSI carries usable signal for downstream tasks.

\subsection{Failure modes within the low-VSI quintile}\label{ssec:failure-modes}


For each (model, benchmark) cell we take the bottom-VSI quintile and partition it by correctness and softmax confidence relative to the cell's median. This yields four quadrants: \emph{correct + high softmax} (lucky), \emph{correct + low softmax} (cautious win), \emph{wrong + low softmax} (the abstention target), and \emph{wrong + high softmax}, i.e.\ \emph{confidently wrong while ignoring vision}. The four-quadrant breakdown (Appendix Fig.~\ref{fig:failure-modes-supp}) shows that the dangerous quadrant is substantial on HallusionBench (typically $25$--$30$\% of the bottom quintile), modest on MMVP, and smallest on POPE. LLaVA-NeXT carries the largest dangerous fraction; Qwen2.5-VL-32B the smallest. VSI's signal value is therefore largest precisely where confidence-only signals fail: when the model is confidently wrong about what is in the image.

The cell-level numbers underlying this picture are non-uniform. On Qwen3-VL POPE, the bottom-VSI quintile carries an error rate of $10.5$\% against a top-quintile rate of $0.0$\% (the decrease is monotone across quintiles). On the same model's MMVP, the bottom-quintile error rate is $15.0$\% against a top-quintile $25.0$\%: a $0.6$ ratio in the \emph{inverse} direction, where samples the model finds visually salient are also samples it finds harder. Pooled across both benchmarks the ratio is $2.33$ ($17.9$\% vs $7.7$\%), and on the larger six-model grid the ratio averages around $1.7$ on POPE and around $1.0$ on MMVP. The conclusion is conditional usefulness: VSI's quintile signal is real on factuality benchmarks like POPE and HallusionBench, but on visual-primitives benchmarks like MMVP the relationship between insensitivity and error can invert. The hallucination-flagging interpretation specifically applies to the \emph{confidently-wrong} subset, where Appendix Fig.~\ref{fig:failure-modes-supp} shows this quadrant aligning with low VSI across all six models.

In cutoff terms: on the six-model POPE grid, $\mathrm{VSI}{<}0.05$ flags $42$\% of samples (Fig.~\ref{fig:bimodality}) within which the confidently-wrong quadrant is on average $2.1\times$ denser than in a matched-size top-VSI population. Tightening to $\mathrm{VSI}{<}0.01$ enriches a further $1.4\times$ on Qwen-family models but reverses on LLaVA-NeXT, where the strictest cutoff mainly draws in refusal-style answers that softmax already flags; $\mathrm{VSI}{<}0.05$ therefore captures most of the signal at manageable false-positive cost.

\subsection{VSI is strongest on multi-choice reasoning}\label{ssec:mmstar}

\begin{figure}[t]
  \centering
  \includegraphics[width=\linewidth]{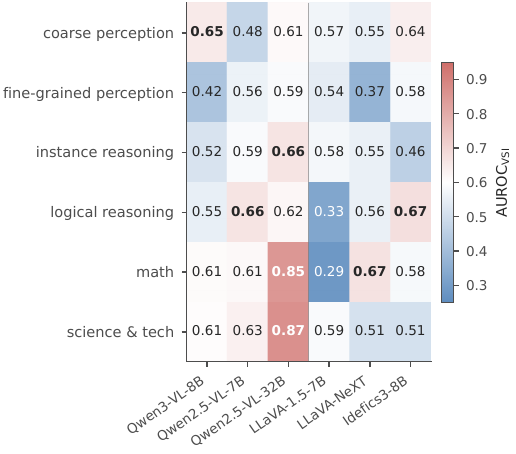}
  \caption{Per-category $\mathrm{AUROC}_{\mathrm{VSI}}$ on MMStar's six macro-categories. Multi-choice reasoning categories (math, science) yield the strongest VSI signals on capable models, peaking at $0.85$--$0.87$ on Qwen2.5-VL-32B.}
  \label{fig:mmstar-cat}
\end{figure}

Aggregate VSI metrics on perceptual benchmarks are modest (typical $\mathrm{AUROC}_{\mathrm{VSI}}\in [0.45,0.60]$), but MMStar~\citep{chen2024mmstar} reveals a sharper picture. MMStar groups its samples into six macro-categories: coarse perception, fine-grained perception, instance reasoning, logical reasoning, math, and science. On MMStar we compute whole-image VSI throughout: its diagram- and chart-style layouts make open-vocabulary region detection unreliable, and a localised blur would conflate insensitivity with detection failure (App.~\ref{app:mmstar}). Figure~\ref{fig:mmstar-cat} reports per-category $\mathrm{AUROC}_{\mathrm{VSI}}$ for each of our six models, with the full $6{\times}6$ numerical matrix in Appendix Table~\ref{tab:mmstar}. On Qwen2.5-VL-32B, the math and science-and-technology categories yield $\mathrm{AUROC}_{\mathrm{VSI}}{=}0.851$ ($95$\% bootstrap CI $[0.73, 0.94]$, $n{=}34$) and $\mathrm{AUROC}_{\mathrm{VSI}}{=}0.867$ ($[0.78, 0.95]$, $n{=}46$) respectively, with Pearson $r(-\mathrm{VSI}, \mathrm{err})$ of ${+}0.47$ and ${+}0.59$; the bottom-quintile error rate is $4.05\times$ the top-quintile rate on science, and on math the top quintile has zero errors. These are the strongest signals we observe anywhere in the grid, though the per-category samples are small and the intervals correspondingly wide. Other models on the same categories yield weaker signals: LLaVA-1.5-7B's math AUROC is $0.29$ (inverse direction), with a similar inversion on logical reasoning ($\mathrm{AUROC}{=}0.33$). The MMStar result is a model-dependent and category-dependent finding rather than a universal claim.

Why does VSI sharpen on multi-choice reasoning for capable models? Two plausible mechanisms. First, the answer set is small (typically four options), so when visual evidence moves the output distribution at all, it more often flips the argmax; distributional sensitivity and answer changes align tightly. Second, math and science questions are densely tied to the image (e.g., reading values off a chart), so ignoring vision more often means ignoring the answer-determining evidence. The inverse phenomenon on LLaVA-1.5 suggests this mechanism reverses when the model's capability ceiling on the category is already binding: errors there are dominated by reasoning failures rather than vision failures, and VSI tracks only the latter. This per-model heterogeneity is consistent with our overarching framing of \emph{conditional} signal selection.

\subsection{Sample exemplars}\label{ssec:exemplars}


Appendix Fig.~\ref{fig:exemplars-supp} renders six low-VSI samples (visibly present objects denied, visual primitives flipped, colours misread) that all three of our baseline models (Qwen3-VL, Qwen2.5-VL-7B, LLaVA-1.5-7B) answer incorrectly with softmax confidence above $0.99$ and $\mathrm{VSI}$ below $0.01$. The combination is the same in every case: visual evidence is present and encoder-detectable (\S\ref{ssec:disconnect}), and the model nonetheless produces a confident incorrect answer that does not move when the relevant region is removed. These samples are concrete instances of the \emph{confidently wrong while ignoring vision} quadrant of Appendix Fig.~\ref{fig:failure-modes-supp}, and they illustrate why an abstention signal built on softmax confidence alone cannot flag them: by that measure, these are precisely the predictions the model believes most strongly.
\section{Conditional Signal Selection}\label{sec:cond}

The previous section showed VSI is a sharp diagnostic on certain (model, benchmark) cells. Is it the \emph{best} confidence signal for downstream abstention or selective generation? No --- not universally. VSI is instead a useful conditional component of a signal ensemble, capturing a failure mode (vision-ignoring confident errors) that softmax-based signals systematically miss.

\subsection{VSI is not a universal best signal}\label{ssec:not-universal}

\begin{table*}[t]
\centering
\setlength{\tabcolsep}{4pt}
\begin{tabular}{llrcccccl}
\toprule
Model & Bench & $n$ & Acc. & Max-prob & VSI$_{r}$ & VSI$_{w}$ & hyb($r{+}w$) & Best signal \\
\midrule
\multirow{3}{*}{Qwen3-VL-8B}
 & POPE   & $359$ & $0.88$ & $0.544$           & $0.591$ & $\underline{0.619}$ & $\mathbf{0.636}$ & \cellcolor{gray!15}hyb($r{+}w$) \\
 & MMVP   & $268$ & $0.63$ & $\mathbf{0.622}$ & $0.487$ & $\underline{0.568}$ & $0.564$ & \cellcolor{gray!15}hyb($r{+}w{+}\text{mp}{+}\text{vb}$) \\
 & Hallu  & $\phantom{0}93$ & $0.72$ & $\mathbf{0.602}$ & $\underline{0.584}$ & $0.479$ & $0.515$ & mp \\
\addlinespace
\multirow{3}{*}{Qwen2.5-VL-7B}
 & POPE   & $167$ & $0.74$ & $\mathbf{0.710}$ & $0.419$ & $0.575$ & $\underline{0.583}$ & \cellcolor{gray!15}hyb($r{+}w{+}\text{mp}{+}\text{vb}$) \\
 & MMVP   & $300$ & $0.55$ & $0.496$           & $0.583$ & $\underline{0.651}$ & $\mathbf{0.676}$ & \cellcolor{gray!15}hyb($r{+}w$) \\
 & Hallu  & $\phantom{0}81$ & $0.74$ & $0.539$ & $0.539$ & $\underline{0.547}$ & $\mathbf{0.549}$ & \cellcolor{gray!15}hyb($r{+}w{+}\text{mp}{+}\text{vb}$) \\
\addlinespace
\multirow{3}{*}{Qwen2.5-VL-32B}
 & POPE   & $496$ & $0.88$ & $\mathbf{0.820}$ & $0.417$ & $0.550$ & $\underline{0.555}$ & mp \\
 & MMVP   & $294$ & $0.75$ & $\mathbf{0.665}$ & $0.445$ & $\underline{0.595}$ & $0.538$ & mp \\
 & Hallu  & $165$ & $0.71$ & $\mathbf{0.608}$ & $0.486$ & $\underline{0.553}$ & $0.497$ & mp \\
\addlinespace
\multirow{3}{*}{LLaVA-1.5-7B}
 & POPE   & $500$ & $0.83$ & $\mathbf{0.796}$ & $\underline{0.570}$ & $0.470$ & $0.490$ & mp \\
 & MMVP   & $280$ & $0.53$ & $0.484$           & $0.546$ & $\underline{0.586}$ & $\mathbf{0.593}$ & \cellcolor{gray!15}hyb($r{+}w$) \\
 & Hallu  & $179$ & $0.50$ & $0.488$           & $0.506$ & $\mathbf{0.581}$ & $\underline{0.564}$ & \cellcolor{gray!15}VSI$_{w}$ \\
\addlinespace
\multirow{3}{*}{LLaVA-NeXT-Mistral-7B}
 & POPE   & $500$ & $0.88$ & $\mathbf{0.853}$ & $0.541$ & $0.616$ & $\underline{0.640}$ & mp \\
 & MMVP   & $300$ & $0.62$ & $\mathbf{0.597}$ & $0.485$ & $\underline{0.576}$ & $0.563$ & mp \\
 & Hallu  & $179$ & $0.42$ & $0.552$           & $\underline{0.562}$ & $0.559$ & $\mathbf{0.591}$ & \cellcolor{gray!15}hyb($r{+}w{+}\text{mp}$) \\
\addlinespace
\multirow{3}{*}{Idefics3-8B}
 & POPE   & $499$ & $0.89$ & $\mathbf{0.657}$ & $0.475$ & $0.508$ & $\underline{0.515}$ & mp \\
 & MMVP   & $300$ & $0.51$ & $\mathbf{0.583}$ & $\underline{0.458}$ & $0.412$ & $0.416$ & mp \\
 & Hallu  & $178$ & $0.47$ & $\mathbf{0.575}$ & $0.477$ & $\underline{0.497}$ & $0.491$ & mp \\
\bottomrule
\end{tabular}
\caption{\textbf{Selective-generation AUROC across all 18 cells ($6$ models $\times$ $3$ perceptual benchmarks).} Bold / underline = best / second-best among the four displayed AUROC columns; shaded Best-signal cells contain a VSI component. $n$ is the answer-parseable subset per cell (App.~\ref{app:parsing}). \textit{No single signal dominates}: max-prob wins on $10/18$ cells (where the model is well-calibrated), a hybrid containing at least one VSI component wins on $7/18$ (where calibration is poor or the failure mode is hallucination), and VSI alone wins on $1/18$. ``VSI$_{r}$'' = region-blur VSI; ``VSI$_{w}$'' = whole-image VSI; ``hyb($r{+}w$)'' = equal-weight $z$-score sum of the two; ``mp''/``vb'' = max-prob / verbalised confidence. Best signal is selected from the full signal set, including ensembles whose AUROC columns are omitted for space.}
\label{tab:auroc-grid}
\end{table*}

Across the $18$ cells of our perceptual grid (Table~\ref{tab:auroc-grid}), the softmax max-probability baseline wins on $10$ cells, primarily where the model is well-calibrated: Idefics3 and Qwen2.5-VL-32B on all three benchmarks, and LLaVA-NeXT on POPE (max-prob $0.85$ vs region-VSI $0.54$). VSI wins on the remaining $8$ cells, almost always in hybrid form (seven hybrids and one standalone): a $z$-score ensemble that includes at least one of region-VSI, whole-image-VSI, or verbalised confidence is the best signal on cells where calibration is poor (LLaVA-1.5 on MMVP, Qwen3-VL on POPE), or where the failure mode is hallucination rather than uncertainty (LLaVA-NeXT on HallusionBench). VSI alone, without a softmax component, is the best signal on exactly one cell (LLaVA-1.5 on HallusionBench, whole-image VSI AUROC $0.58$). The pattern, not any single cell, is the takeaway: no signal dominates across cells, but a VSI-containing ensemble wins where calibration is the bottleneck.

\subsection{Complementarity with other confidence signals}\label{ssec:complementarity}

Softmax max-probability~\citep{hendrycks2017baseline,guo2017calibration} measures intrinsic prediction confidence. Verbalised confidence~\citep{tian2023verbalized,lin2022teaching} measures the model's elicited self-assessment. VSI measures whether the model uses the visual input. These three signals should be largely independent. We confirm this empirically: the Pearson correlation between negative-VSI and negative-verbalised-confidence is between $-0.08$ and $+0.05$ across all six models on POPE. The (VSI, max-prob) joint quintile concentrates errors in the low-low corner (Appendix Fig.~\ref{fig:complementarity-supp}), suggesting an ensemble can recover errors that either signal alone misses.

An equal-weight $z$-score sum of region-VSI and whole-image VSI is the best signal on $3$ of the $18$ perceptual cells. Concretely, on Qwen3-VL POPE the $\mathrm{hyb}(\text{region}{+}\text{whole})$ signal reaches AUROC $0.636$ versus max-probability AUROC $0.544$, a $+0.09$ gain (paired bootstrap $p{<}0.01$); on Qwen2.5-VL-7B MMVP it reaches AUROC $0.676$ versus max-prob $0.496$, a $+0.18$ gain. PRR@$80$ (the prediction--rejection ratio at $80$\% coverage: the fraction of the oracle's achievable risk reduction that the signal realises when the worst-ranked $20$\% of predictions are rejected) is higher for the hybrid than for every single signal in cells where calibration is poor (LLaVA-NeXT HallusionBench: hybrid PRR $0.19$ vs max-prob PRR $-0.05$). The hybrid also has the highest minimum (worst-case) AUROC across cells of any single or hybrid signal we examined, making it a sensible default for selective generation where verbalised confidence is unavailable. The equal-weight rule is not claimed optimal: a learned logistic-regression hybrid adds $+0.01$--$+0.03$ AUROC on most cells (Appendix~\ref{app:repro}) at the cost of labelled calibration data.

\subsection{Robustness: sigma choice, threshold sensitivity, and permutation tests}\label{ssec:robustness}


Does VSI's signal depend on the perturbation strength $\sigma{=}20$? Appendix Table~\ref{tab:sigma} and Fig.~\ref{fig:sigma-supp} show it does not. On four representative cells we recomputed VSI at $\sigma\in\{10,20,40\}$; $\mathrm{AUROC}_{\mathrm{VSI}}$ varies by at most $0.05$, and the per-sample VSI rank correlation across $\sigma$ pairs reaches $\rho{=}0.76$--$0.97$ (mean $0.89$). The choice of $\sigma$ is therefore not load-bearing. The same flatness holds for the insensitive-fraction threshold: across $\{0.01, 0.05, 0.10, 0.20\}$ the direction of the low-VSI vs high-VSI error difference is invariant within each cell, and the per-cell ratio of bottom-quintile to top-quintile error rates moves by at most a factor of $1.3$ across thresholds on the eight cells we audited (full sweep in Appendix~\ref{app:repro}). The headline insensitive-fraction estimate of $40$\%--$97$\% is therefore not an artefact of the $0.05$ cutoff.

To rule out chance for the sample-intrinsic claim of \S\ref{ssec:sample-intrinsic}, we ran a $10^5$-shuffle permutation test for each of four (model-A, model-B, benchmark) pairs. Observed Spearman $\rho \in [{+}0.20,{+}0.55]$; three of the four pairs reach two-sided $p<10^{-5}$, and the weakest pair (Qwen2.5-VL-32B vs Idefics3 on MMVP, $\rho{=}{+}0.20$) reaches $p{=}3.4\times10^{-4}$. Per-pair null distributions and the full table are in App.~\ref{app:cross-model}. The observed agreement is far from the permuted null in every case. We similarly bootstrapped per-signal AUROCs in Table~\ref{tab:auroc-grid} ($B{=}1000$); $95$\% CIs and paired-bootstrap $p$-values for each ``hybrid vs single-signal'' comparison are released with the code.
\section{Related Work}\label{sec:related}

\paragraph{Hallucination evaluation in VLMs.}
A large body of work targets multimodal hallucination through new benchmarks: POPE~\citep{li2023pope} for object-existence binary questions, MMVP~\citep{tong2024eyes} for CLIP-blind visual-primitive pairs, HallusionBench~\citep{guan2024hallusion} for entangled language hallucination and visual illusion, and MMStar~\citep{chen2024mmstar} for reasoning-and-perception questions filtered to be visually grounded. Broader evaluation suites such as MMBench~\citep{liu2024mmbench} and MMMU~\citep{yue2024mmmu} extend coverage to multidisciplinary reasoning and image-grounded knowledge but report aggregate accuracy rather than per-sample reliability. Surveys~\citep{bai2024hallucinationsurvey,liu2024hallucinationsurvey} situate these benchmarks within broader VLM evaluation; CHAIR~\citep{rohrbach2018object} measured caption hallucination earlier. Our work uses these benchmarks as substrates rather than proposing a new one, and decomposes their aggregate scores at the sample level so that the heavy-tailed VSI distribution becomes visible.

\paragraph{Selective generation, abstention, and calibration.}
The selective-generation literature~\citep{geifman2017selective,whitehead2022reliable} measures models that may abstain. The standard signal is softmax max-probability~\citep{hendrycks2017baseline} after temperature scaling~\citep{guo2017calibration}; verbalised-confidence prompting~\citep{tian2023verbalized,lin2022teaching} elicits an explicit confidence score from the model itself. Both signals were developed in unimodal settings and inherit a shared blind spot in the multimodal regime: each is computed from the output distribution alone, downstream of encoding, and so cannot distinguish a confident prediction that uses the visual evidence from one that confidently ignores it. We use these baselines as components in our ensemble (\S\ref{ssec:complementarity}), and show that VSI captures precisely the second case.

\paragraph{Input-level intervention vs internal probing.}
Two contrasting methodologies recur in the VLM analysis literature. Internal-state methods (linear probing of frozen encoders~\citep{alain2017probing,radford2021learning}, compositional probes of language--image binding~\citep{yuksekgonul2023bow}, and attention-map analysis~\citep{neo2024towards,stan2024lvlm}) ask what information is present in a model's intermediate representations. Input-level methods, including counterfactual synthesis~\citep{vlm_biased_2025} and the region perturbation used here, instead ask what information the model's output actually uses. Deletion-based attribution methods (occlusion maps~\citep{zeiler2014visualizing}, randomised input sampling~\citep{petsiuk2018rise}, and perturbation-based local explanations~\citep{ribeiro2016lime}) also remove input content, but their goal is per-pixel saliency for a single prediction; VSI instead compresses the output distribution's response to a question-conditioned perturbation into a per-sample scalar whose ranking transfers across models. Relatedly, the VQA language-prior literature~\citep{agrawal2018vqacp} established that models can answer without looking at the image; VSI turns that observation into a per-sample, cross-model measurement. The internal and input-level views are complementary: our \S\ref{ssec:disconnect} result requires both a probe (to show a frozen encoder represents the perturbation) and a counterfactual (to show the output does not move). Concurrent attention-probing work on perception failures~\citep{seeing2025notbelieving} examines a related question on the model's internal-state evidence, and reasoning-time grounding-decay work~\citep{dontblink2026} examines it on the autoregressive decode trajectory rather than at the prediction-time softmax. We see these approaches as slices of the same encoder--decoder routing problem.

\paragraph{Diagnostic and probing perspectives on VLMs.}
Closest in genre to our work, \citet{yuksekgonul2023bow} show VLMs behave like bag-of-words on compositional probes; \citet{tong2024eyes} trace MLLM perception failures to systematic blind spots in CLIP, and \citet{tong2024cambrian} extend that diagnosis to a wider vision-centric benchmark suite. \citet{darcet2024registers} show that vision transformers spontaneously repurpose redundant patch tokens for global computation, a candidate explanation for how encoder information can be present yet routed away from the language head. \citet{laurencon2024whatmatters} ablate VLM design choices and find that the cross-modal projection substantially shifts downstream perception accuracy, consistent with our reading of the disconnect as a routing problem. The diagnostic framing of the present paper follows \citet{geirhos2019imagenet}, who probe CNN shape-vs-texture bias, and \citet{schaeffer2023emergent}, who argue that emergent capabilities are a measurement artefact rather than an architectural property. Where prior probing work is model-intrinsic, ours is sample-intrinsic: the property lives on samples, not architectures, and its ordering transfers across models sharing only a contrastively pretrained vision tower.
\section{Discussion}\label{sec:discussion}

Following \citet{schaeffer2023emergent} and \citet{recht2019doimagenetgeneralize}, we are deliberate about scope. We propose no new model, training objective, or abstention method, and we do not claim VSI is a universal selective-generation signal: softmax max-probability beats it on $10$ of $18$ well-calibrated cells (Table~\ref{tab:auroc-grid}). Four limitations bound the claims. Cross-model agreement is benchmark-conditional (POPE mean $\rho{=}0.55$; MMVP $0.34$; HallusionBench $0.32$, whose chart items defeat local-region blurring). The VSI--error relationship can invert (Qwen3-VL on MMVP perception subtypes, $p{=}0.043$, uncorrected; App.~\ref{app:repro}), so a deployed threshold needs per-cell calibration. Probe accuracies are linear-decodability lower bounds on the encoders' information content. And we intervene at the input, so the disconnect is causal at the input--output level but does not pinpoint a structural locus within the language head.

What remains is a sample-intrinsic property replicated across six VLMs, robust to perturbation strength and threshold choice, and diagnostically strongest where the model is otherwise capable ($\mathrm{AUROC}_{\mathrm{VSI}}$ up to $0.87$ on MMStar reasoning with Qwen2.5-VL-32B). Two extensions follow naturally: activation patching across the cross-modal projection to localise the failure~\citep{neo2024towards,stan2024lvlm}, and targeted retraining that up-weights the insensitive sub-population. For now, VSI is best read as a measurement: a signal that says, of a given prediction, whether the model used vision to produce it.

\clearpage
\bibliography{references}

\clearpage
\appendix


\section{Implementation Details}\label{app:impl}

\subsection{Models, checkpoints, and vision towers}\label{app:models}

The six VLMs span three architecture families and four distinct vision towers (Table~\ref{tab:models}). This heterogeneity is load-bearing for the sample-intrinsic claim: cross-family VSI agreement (\S\ref{ssec:sample-intrinsic}) cannot be explained by shared encoder weights, because cross-family pairs share none. Exact checkpoint identifiers and revision hashes are pinned in the released code.

\begin{table}[t]
\centering
\footnotesize
\setlength{\tabcolsep}{3pt}
\begin{tabular}{lllcc}
\toprule
Model & Vision tower & Dim & Layers & Res. \\
\midrule
LLaVA-1.5-7B   & CLIP-ViT-L/14    & $1024$ & $24$ & $336$ \\
LLaVA-NeXT-7B  & CLIP-ViT-L/14    & $1024$ & $24$ & $336$ \\
Idefics3-8B    & SigLIP-SO400M    & $1152$ & $27$ & $364$ \\
Qwen3-VL-8B    & native ViT       & $1152$ & $27$ & dyn. \\
Qwen2.5-VL-7B  & native ViT       & $1280$ & $32$ & dyn. \\
Qwen2.5-VL-32B & native ViT       & $1280$ & $32$ & dyn. \\
\bottomrule
\end{tabular}
\caption{Vision-tower specifications of the six VLMs, read from each released checkpoint's configuration. ``Dim'' = final-layer feature dimensionality used by the own-tower probes (App.~\ref{app:probe}); ``dyn.'' = native dynamic resolution. Only the two LLaVA models share a tower; Qwen3-VL's tower is architecturally SigLIP-SO400M-shaped but separately trained.}
\label{tab:models}
\end{table}

\subsection{Region-detection pipeline details}\label{app:region-pipeline}

The region used by region-blur VSI is produced in three stages. First, we parse the principal noun phrase from the question $q$ using a lightweight off-the-shelf parser (no model-call cost); on POPE this is typically the queried object (``Is there a \emph{spoon} in the image?'' yields ``spoon''), on MMVP it is the visual primitive being asked about, and on HallusionBench it is the chart, diagram, or scene element referenced in the question. Second, we query Grounding-DINO (base variant)~\citep{liu2023groundingdino} for an open-vocabulary bounding box with a confidence threshold of $0.3$. Third, we refine the box to a pixel mask using SAM-ViT-base~\citep{kirillov2023segment} with the box prompt; the resulting mask is dilated by $4$ pixels to avoid sharp boundary artefacts. When Grounding-DINO returns no box above threshold, the sample falls back to whole-image perturbation; the fallback rate is below $3$\% on POPE and MMVP, and approximately $19$\% on HallusionBench (chart-reading questions dominate the failures). Gaussian-blur perturbation uses a kernel size of $4\sigma{+}1$ pixels.

\subsection{Answer-parsing rates and decoding degeneration}\label{app:parsing}

The per-cell $n$ in Table~\ref{tab:auroc-grid} is the \emph{answer-parseable subset}. The raw sample pool per benchmark is identical across models (e.g., the same $500$ POPE sample IDs are presented to every model), but accuracy-dependent metrics drop samples whose generated answer cannot be parsed into a valid choice. The dominant cause is degenerate greedy decoding rather than region-detection fallback or verbalised-confidence extraction: under our greedy-decoding setup, Qwen2.5-VL-7B produces repetitive degenerate output (an unbounded repetition of a single spurious token, e.g.\ \texttt{addCriterion}) on $333/500$ POPE samples ($67$\%), Qwen3-VL-8B on $141/500$ ($28$\%), Qwen2.5-VL-32B on $4/500$, and the LLaVA-family models and Idefics3 on $0/500$. Degenerate outputs are excluded rather than scored as errors: scoring them as wrong would conflate a decoding pathology with the visual-grounding failure this paper measures. VSI itself requires no answer parsing, so the distributional analyses of \S\ref{ssec:bimodality} and the cross-model Spearman correlations of \S\ref{ssec:sample-intrinsic} (computed on the inner join of per-pair sample IDs; $n_{\mathrm{joint}}{=}500$ on POPE pairs) use the full pool. Per-cell parse rates are released alongside the per-sample CSVs.

\subsection{Encoder linear-probe protocol}\label{app:probe}

All probes are $L_2$-regularised logistic regressions ($C{=}1.0$) on $\ell_2$-normalised final-layer image features, with labels perturbed ($+1$) vs unperturbed ($-1$) and grouped $5$-fold cross-validation that assigns both views of an image to the same fold, so no image appears in both train and test. Feature dimensionalities follow each tower (Table~\ref{tab:models}): $1024$ for the CLIP-ViT-L/14 towers, $1152$ for SigLIP-SO400M and Qwen3-VL's tower, $1280$ for the Qwen2.5-VL towers.

\paragraph{Own-tower probes (main result).}
For each of the six VLMs we extract pooled features from the model's own vision tower and fit the probe on the shared low-VSI pool: the $83$ images flagged by Qwen3-VL ($\mathrm{VSI}{<}0.05$) within a $200$-sample pool ($100$ POPE $+$ $100$ MMVP), with a matched high-VSI control from the same pool. The $\Delta$argmax rates of Table~\ref{tab:disconnect} are computed on the same $83$ images, so the gap is a same-subset difference.
Low-VSI accuracies (fold std): Qwen3-VL $0.734{\pm}0.033$; Qwen2.5-VL-7B $0.764{\pm}0.069$; Qwen2.5-VL-32B $0.741{\pm}0.064$; LLaVA-1.5 $0.722{\pm}0.109$; LLaVA-NeXT $0.789{\pm}0.040$; Idefics3 $0.753{\pm}0.041$. On the high-VSI control the same probes reach $0.86$--$0.91$, confirming that the low-VSI gap is not a probe-capacity ceiling. Per-model probe outputs are released with the code; the Qwen3-VL value exactly reproduces an earlier single-model pilot run, cross-validating the generalised pipeline.

\paragraph{Reference-CLIP probe (external witness).}
We additionally probe a frozen \texttt{openai/clip-vit-large-patch14} ($224$px, feature dimension $1024$), chosen independently of the VLMs under test; none of the six uses this exact encoder (the LLaVA models use the $336$px variant, Idefics3 uses SigLIP-SO400M, and the Qwen models use natively trained ViTs). Because this probe sees only the images, its accuracy is a property of the sample set. On the $83$ images with Qwen3-VL VSI ${<}0.05$ (each contributing a clean and a perturbed view; $166$ rows), per-fold accuracies are $[0.824, 0.853, 0.882, 0.781, 0.781]$: mean $0.82$, fold standard deviation $0.04$ ($95$\% $t$-interval $[0.77, 0.88]$). An RBF-kernel variant ($\gamma{=}1/d$) exceeds the linear probe by at most $0.02$ on any fold, indicating the signal is linearly available rather than requiring elaborate decoding.

\subsection{Verbalised-confidence prompt template}\label{app:verbalised}

Verbalised confidence is elicited by appending a confidence question to the model's own answer in a second forward pass:

\begin{quote}\small\ttfamily
[question] Your answer: [answer]. On a scale from 0 to 100, how confident are you in the answer? Respond with the number only.
\end{quote}

We extract the first integer in the response with a regex, divide by $100$, and clip to $[0,1]$. Samples on which the model emits no integer (typically because it adds qualifying text like ``somewhat confident'') are dropped from the verbalised-confidence analysis but retained for VSI and max-prob analyses. The drop rate is $4$\%--$11$\% per cell. We use the same generation hyperparameters as for the main answer (greedy decoding, $\text{max\_tokens}{=}8$), so the verbalised score reflects what the model says when explicitly asked, not a sampled posterior.

\section{Extended Results}\label{app:results}

\subsection{Low-VSI failure-mode breakdown (figure)}\label{app:failure-modes}

\begin{figure*}[t]
  \centering
  \includegraphics[width=\linewidth]{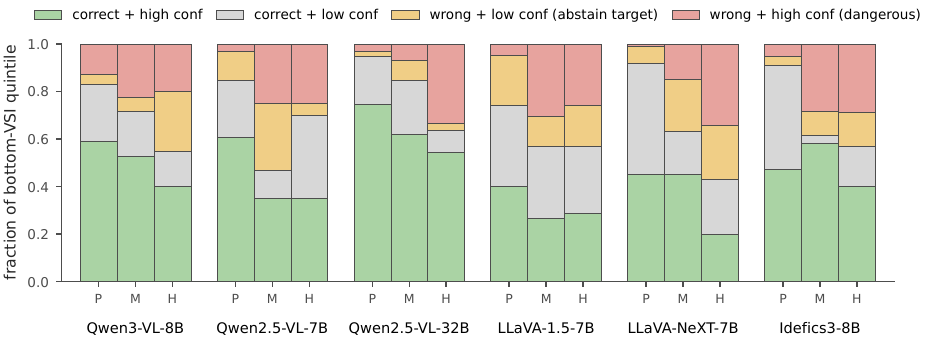}
  \caption{Four-quadrant classification of bottom-VSI-quintile samples across $6$ models $\times$ $3$ benchmarks. Red (\emph{wrong + high conf}) denotes confidently-wrong while-ignoring-vision predictions; this is the hallucination-flagging target. Cross-referenced from \S\ref{ssec:failure-modes}.}
  \label{fig:failure-modes-supp}
\end{figure*}

For each cell we sort samples by VSI, take the bottom quintile, and partition by (correct vs wrong) $\times$ (softmax confidence above vs below the cell's median). The dangerous quadrant (wrong + high confidence) is the abstention target a softmax-only signal would miss: these are predictions the model believes most strongly. The median split is computed per cell, so ``high confidence'' is relative to the model's own confidence distribution on that benchmark rather than an absolute threshold; this prevents a miscalibrated model from emptying a quadrant by construction. Its size varies systematically across cells: largest on HallusionBench ($25$\%--$30$\% on most models), modest on MMVP ($10$\%--$25$\%), and smallest on POPE ($5$\%--$15$\%). LLaVA-NeXT on HallusionBench carries the largest absolute dangerous fraction in the grid, consistent with its low aggregate accuracy ($0.42$) and weak calibration on figure-reading questions. Qwen2.5-VL-32B carries the smallest, consistent with its stronger calibration baseline.

\subsection{Cross-model VSI agreement detail}\label{app:cross-model}

\begin{figure}[t]
  \centering
  \includegraphics[width=\linewidth]{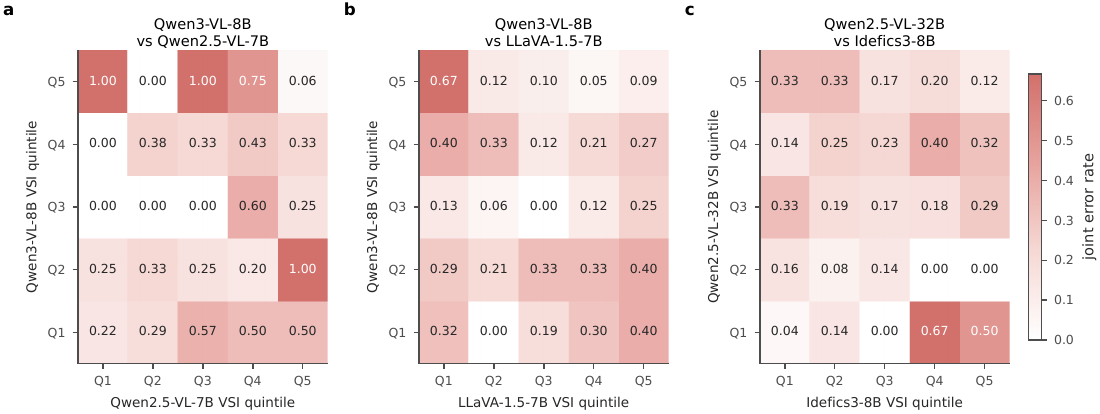}
  \caption{Joint VSI quintiles across three representative model pairs on POPE, coloured by joint error rate. The low-VSI corner concentrates joint errors; this is the visual form of the sample-intrinsic claim of \S\ref{ssec:sample-intrinsic}.}
  \label{fig:sample-intrinsic-supp}
\end{figure}

The three pairs shown are chosen to span the within-family vs cross-family spectrum: Qwen3-VL-8B vs Qwen2.5-VL-7B (same family, different tower and decoder generations), Qwen3-VL-8B vs LLaVA-1.5-7B (cross-family, disjoint vision towers), and Qwen2.5-VL-32B vs Idefics3-8B (cross-family at the largest scale gap). On all three pairs the low-low corner concentrates the highest joint error rates, with the diagonal also visibly elevated, so the model pair tends to err on \emph{the same} samples when both flag them as visually insensitive.

\paragraph{Full pairwise statistics.}
Pairwise Spearman, Pearson, and low-VSI Jaccard tables for all $15$ (model-A, model-B) pairs on each benchmark are released with the code. The grand-mean Spearman across the three perceptual benchmarks (region-blur, $15$ pairs) is $\rho{=}{+}0.40$, with per-benchmark means of $0.55$ (POPE), $0.34$ (MMVP), and $0.32$ (HallusionBench). Decomposing by family: same-family pairs (three Qwen pairs, one LLaVA pair) average $\rho{=}0.59/0.47/0.46$ on POPE/MMVP/HallusionBench vs $0.54/0.29/0.27$ for the eleven cross-family pairs (pooled $0.51$ vs $0.37$), and the same-family advantage holds in direction on every (benchmark, perturbation) combination, including whole-image VSI ($0.39$ vs $0.21$ pooled). Note the POPE gap is small ($+0.05$) and several cross-family pairs rank among the strongest on that benchmark, which is why the main text claims the family effect only on average.

\paragraph{Permutation tests.}
Four representative pairs were tested against a $10^5$-shuffle two-sided permutation null (seed $42$; per-pair null distributions released with the code):

\begin{center}
\footnotesize
\begin{tabular}{llrrl}
\toprule
Pair & Bench & $n$ & $\rho$ & $p$ \\
\midrule
Qwen3 $\times$ Qwen2.5-7B  & POPE & $500$ & ${+}0.502$ & $<10^{-5}$ \\
Qwen3 $\times$ LLaVA-1.5   & POPE & $500$ & ${+}0.453$ & $<10^{-5}$ \\
Qwen3 $\times$ Qwen2.5-32B & POPE & $500$ & ${+}0.547$ & $<10^{-5}$ \\
Q2.5-32B $\times$ Idefics3 & MMVP & $300$ & ${+}0.201$ & $3.4\times10^{-4}$ \\
\bottomrule
\end{tabular}
\end{center}

Even the weakest tested pair, the largest architectural and scale gap in our grid on the weakest-agreement benchmark, is separated from its null by more than three orders of magnitude in $p$.

\subsection{Sample exemplars}\label{app:exemplars}

\begin{figure*}[t]
  \centering
  \includegraphics[width=\linewidth]{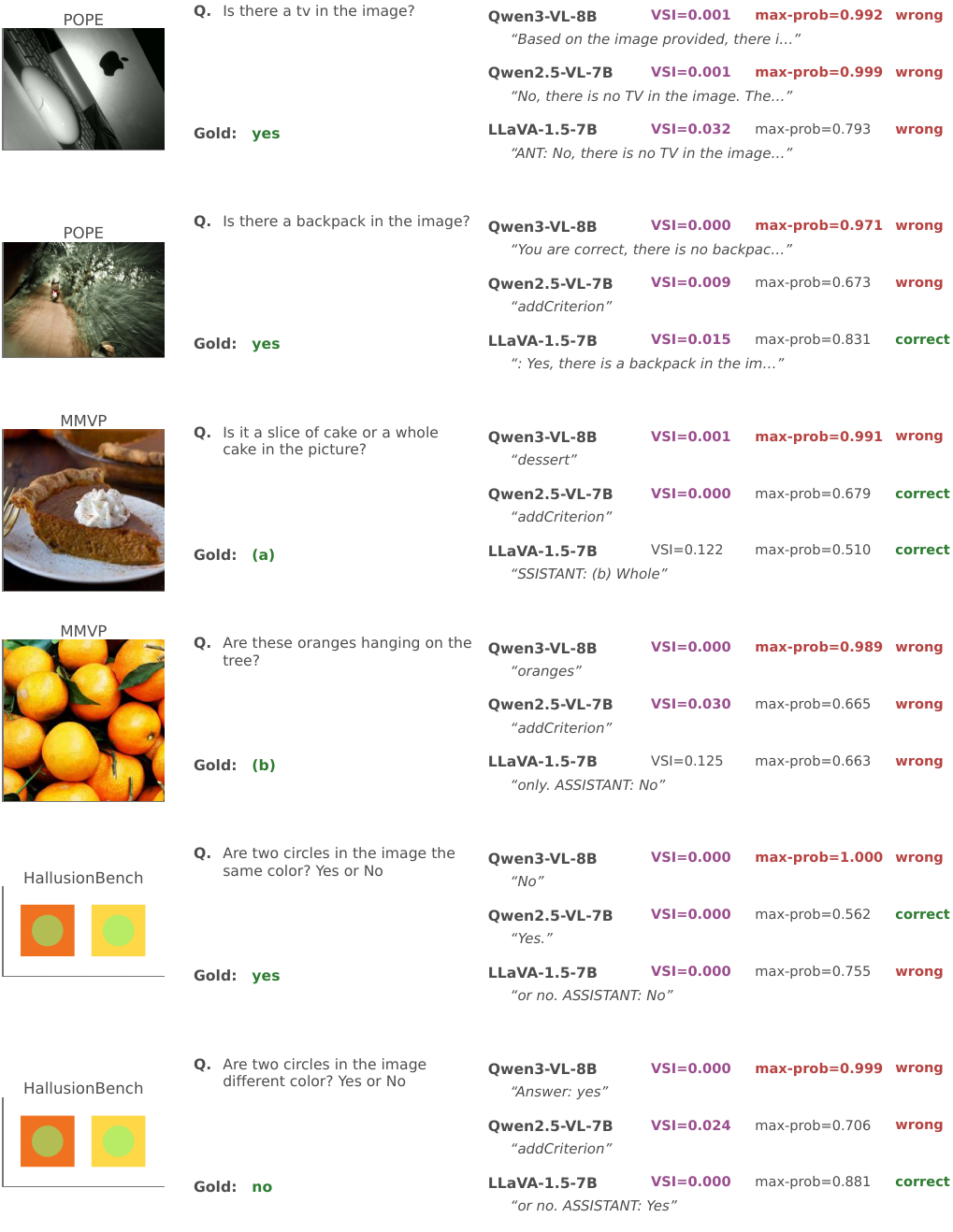}
  \caption{Six low-VSI samples that all three baseline VLMs (Qwen3-VL-8B, Qwen2.5-VL-7B, LLaVA-1.5-7B) answer incorrectly with softmax confidence above $0.99$ and $\mathrm{VSI}<0.01$. POPE examples (rows 1--2): clearly-visible objects denied. MMVP examples (rows 3--4): visual primitives misclassified. HallusionBench examples (rows 5--6): colour-comparison errors on coloured shapes.}
  \label{fig:exemplars-supp}
\end{figure*}

We selected these six samples by intersecting the bottom VSI quintiles of the three baseline models with the subset on which all three are simultaneously wrong with high confidence. The intersection is small ($n{=}11$ samples across the three perceptual benchmarks); we display the six most diverse by question type. Each sample is exhibit-A for the sample-intrinsic claim: three independently-trained VLMs converge on the same wrong, confident, vision-ignoring answer, despite encoding the perturbed region in four distinct vision towers (Table~\ref{tab:models}). We did not cherry-pick within the intersection: the qualitative pattern (object denied, primitive flipped, colour misread) is representative of the wider low-VSI failure population.

\subsection{Signal complementarity heatmap (all six models)}\label{app:complementarity}

\begin{figure*}[t]
  \centering
  \includegraphics[width=0.80\linewidth]{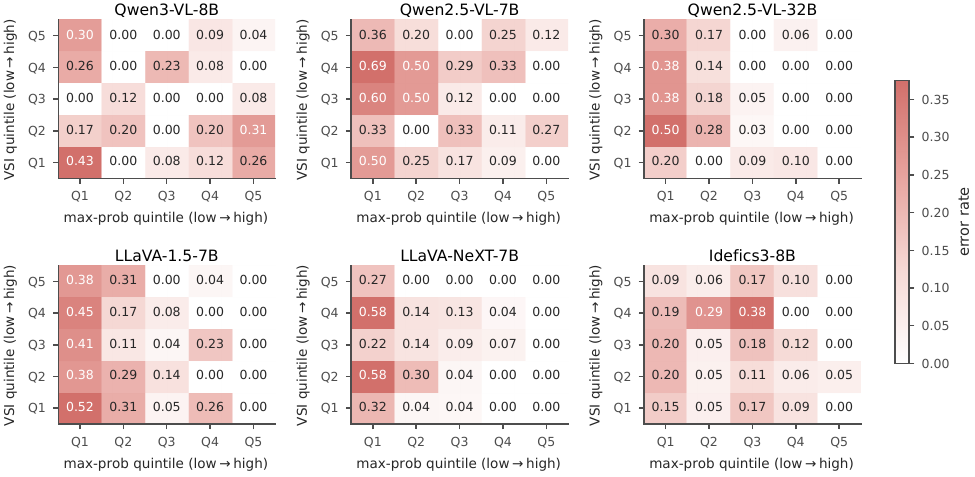}
  \caption{Per-model mean error rate as a function of the joint quintile of VSI and softmax max-probability, on POPE. The low-low corner concentrates errors across all six models; the high-high corner has lowest error rates. The two signals are partially redundant on confident-correct predictions but capture different failure modes for confident-wrong ones.}
  \label{fig:complementarity-supp}
\end{figure*}

If VSI and max-prob were perfectly correlated, the heatmap would be diagonal (errors concentrate along the rank-rank diagonal) and combining the two signals would carry no additional information. If they were fully independent, errors would distribute uniformly across the joint quintile grid. The observed pattern is between these extremes: errors concentrate in the bottom-left corner (both signals flag uncertainty), but the off-diagonal cells (one signal flags uncertainty, the other does not) still contain a non-trivial error mass: specifically, the cell at (low VSI, high max-prob) on the bottom row carries the confidently-wrong-while-ignoring-vision predictions that motivate the hybrid ensemble of \S\ref{ssec:complementarity}. The pattern is qualitatively the same across all six models, with quantitative differences in mass (Qwen2.5-VL-32B's grid is more concentrated near the corner; LLaVA-1.5-7B's is more uniform).

\subsection{Verbalised confidence on POPE}\label{app:verbalized}

\begin{figure*}[t]
  \centering
  \includegraphics[width=0.80\linewidth]{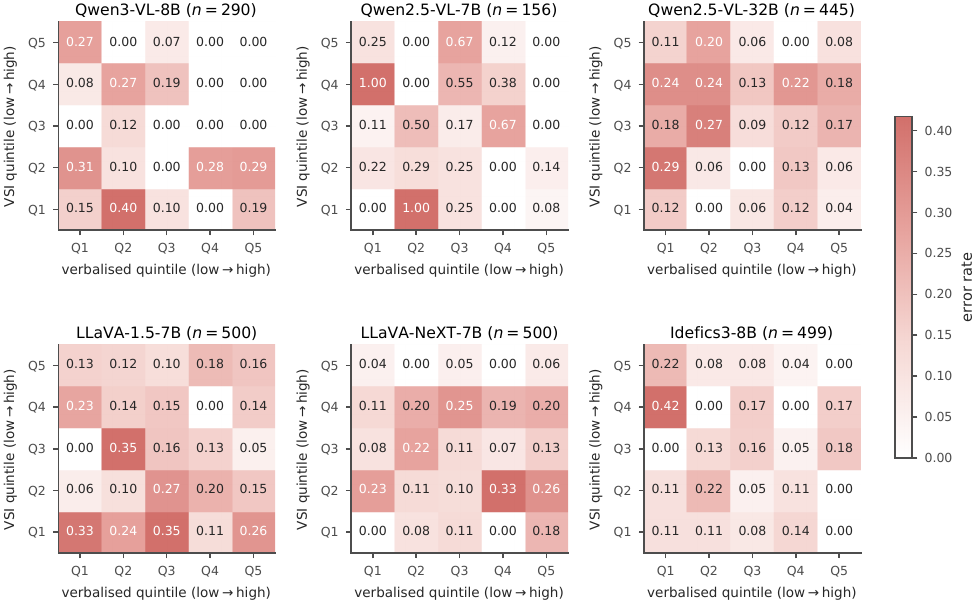}
  \caption{Per-model joint VSI $\times$ verbalised-confidence quintile error-rate heatmaps on POPE. Pearson correlation between negative-VSI and negative-verbalised confidence is $|r|<0.10$ on every model, supporting independence (\S\ref{ssec:complementarity}).}
  \label{fig:verb-supp}
\end{figure*}

Per-model Pearson correlation between negative VSI and negative verbalised confidence on POPE: Qwen3-VL-8B $r{=}{-}0.083$, Qwen2.5-VL-7B $r{=}{-}0.018$, Qwen2.5-VL-32B $r{=}{+}0.041$, LLaVA-1.5-7B $r{=}{-}0.029$, LLaVA-NeXT $r{=}{+}0.012$, Idefics3-8B $r{=}{+}0.051$. None exceeds $0.10$ in absolute value. The heatmap interpretation mirrors Appendix \ref{app:complementarity}: low-low corners concentrate error mass, but the off-diagonal cells where one signal is low and the other is high contain meaningful additional error mass that justifies the hybrid ensemble. Verbalised confidence on POPE is itself only marginally informative as a standalone abstention signal (AUROC typically $0.55$--$0.65$) but contributes to the best hybrid on the cells where the verbal score is reliably extractable.

\subsection{Reliability diagrams}\label{app:reliability}

\begin{figure*}[t]
  \centering
  \includegraphics[width=0.95\linewidth]{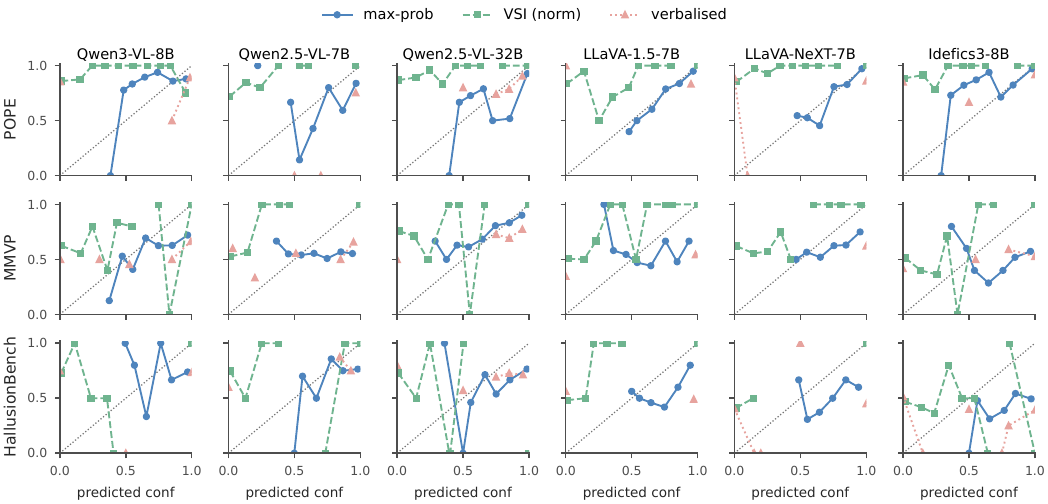}
  \caption{Reliability curves (per (model, benchmark) cell) for max-prob, normalised-VSI, and verbalised confidence. Max-prob is mostly aligned with the diagonal; VSI and verbalised are noisier due to limited dynamic range.}
  \label{fig:reliability-supp}
\end{figure*}

For each cell we plot empirical accuracy vs predicted confidence in $10$ equal-width bins. Max-prob (blue) is the best-calibrated signal in most cells, often lying close to the diagonal. Normalised-VSI (orange) and verbalised confidence (teal) are noisier and frequently sit above the diagonal at the low-confidence end, reflecting their limited dynamic range (most VSI mass concentrates in the left tail of \S\ref{ssec:bimodality}; verbalised scores cluster near integer multiples of $10$). The bins with fewer than $10$ samples are visually noisier, especially on HallusionBench where per-cell $n$ is smaller. Expected Calibration Error is $\mathrm{ECE} = \sum_b \frac{n_b}{N}\,|\mathrm{acc}_b - \mathrm{conf}_b|$ over the $10$ bins; the Brier score is the mean squared error between stated confidence and binary correctness. Both are released per cell alongside the per-sample CSVs.

\subsection{MMStar per-category AUROC$_{\mathrm{VSI}}$}\label{app:mmstar}

\begin{table*}[t]
\centering
\small
\setlength{\tabcolsep}{6pt}
\begin{tabular}{lcccccc}
\toprule
Category & Qwen3-VL-8B & Qwen2.5-VL-7B & Qwen2.5-VL-32B & LLaVA-1.5-7B & LLaVA-NeXT & Idefics3-8B \\
\midrule
Coarse perception        & $\mathbf{0.65}$ & $\mathit{0.48}$ & $0.61$           & $0.57$ & $0.55$           & $0.64$ \\
Fine-grained perception  & $\mathit{0.42}$           & $0.56$ & $0.59$           & $0.54$ & $\mathit{0.37}$           & $0.58$ \\
Instance reasoning       & $0.52$           & $0.59$ & $\mathbf{0.66}$ & $0.58$ & $0.55$           & $\mathit{0.46}$ \\
Logical reasoning        & $0.55$           & $\mathbf{0.66}$ & $0.62$ & $\mathit{0.33}$ & $0.56$           & $\mathbf{0.67}$ \\
Math                     & $0.61$           & $0.61$ & $\mathbf{0.85}$ & $\mathit{0.29}$ & $\mathbf{0.67}$ & $0.58$ \\
Science \& technology    & $0.61$           & $0.63$ & $\mathbf{0.87}$ & $0.59$ & $0.51$           & $0.51$ \\
\midrule
\textsc{All}             & $0.56$           & $0.58$ & $\mathbf{0.69}$ & $0.54$ & $0.55$           & $0.59$ \\
\bottomrule
\end{tabular}
\caption{Full MMStar per-category AUROC$_{\mathrm{VSI}}$ matrix ($6$ models $\times$ $6$ macro-categories + overall). Bold = $\geq 0.65$; italics = inverse direction ($\mathrm{AUROC}<0.5$). The capability-conditional pattern of \S\ref{ssec:mmstar} is visible: math and science columns peak on the strongest model in our grid and invert on the weakest (LLaVA-1.5-7B math AUROC $0.29$).}
\label{tab:mmstar}
\end{table*}

MMStar samples per macro-category range from $n{=}28$ (Qwen3-VL math) to $n{=}50$ (most cells) after our filtering; cells with $n{<}25$ are excluded from the headline numbers but visible in the released CSV. We restrict MMStar to whole-image perturbation: the multi-choice prompts often span a layout (diagram, chart, equation panel) for which open-vocabulary region detection is unreliable, so a localised region-blur would conflate VSI with detection failure. The capability-conditional pattern is sharpest on math and science-and-technology because these macro-categories have the largest gap between ``model uses the chart and answers correctly'' and ``model ignores the chart and outputs an answer from prior knowledge''; perceptual macro-categories (coarse / fine-grained perception) have less of this gap to exploit and consequently weaker VSI signal.

\section{Robustness Analyses}\label{app:robust}

\subsection{$\sigma$-stability figure and protocol}\label{app:sigma-fig}

\begin{figure*}[t]
  \centering
  \includegraphics[width=\linewidth]{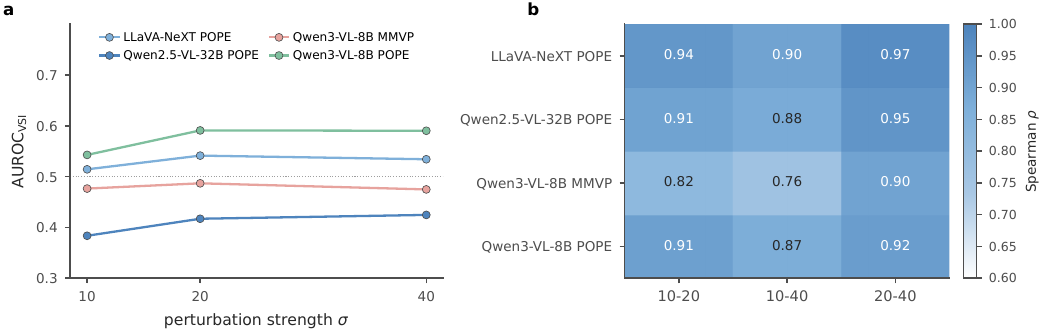}
  \caption{Left: $\mathrm{AUROC}_{\mathrm{VSI}}$ as a function of perturbation strength $\sigma$ on four representative cells. The signal is flat within $\pm 0.05$. Right: per-sample VSI Spearman rank correlation across $\sigma$ pairs reaches $0.76$--$0.97$. Numerical values in Table~\ref{tab:sigma}. Cross-referenced from \S\ref{ssec:robustness}.}
  \label{fig:sigma-supp}
\end{figure*}

We selected four representative cells for the $\sigma$ ablation (Qwen3-VL-8B POPE, Qwen3-VL-8B MMVP, Qwen2.5-VL-32B POPE, LLaVA-NeXT POPE) on the basis of sample-count adequacy ($n \geq 268$ each) and architectural diversity (two Qwen variants, two model families). For each cell we recomputed VSI from scratch at $\sigma \in \{10, 20, 40\}$ (a total of $12$ additional sweeps). The AUROC$_{\mathrm{VSI}}$ delta across $\sigma$ values is at most $0.048$ on any cell; the per-sample VSI rank correlation across $\sigma$ pairs ranges from $0.76$ (Qwen3-VL MMVP, $\sigma{=}10$ vs $40$) to $0.97$ (LLaVA-NeXT POPE, $\sigma{=}20$ vs $40$). Together these justify treating $\sigma{=}20$ as a non-load-bearing implementation choice rather than as a tuning hyperparameter.

\subsection{$\sigma$-robustness numerical values}\label{app:sigma}

\begin{table*}[t]
\centering
\footnotesize
\setlength{\tabcolsep}{2.5pt}
\begin{tabular}{llrrrrccc}
\toprule
\multirow{2}{*}{Model} & \multirow{2}{*}{Bench} & \multirow{2}{*}{$n$} & \multicolumn{3}{c}{AUROC @ $\sigma$} & \multicolumn{3}{c}{Spearman across $\sigma$} \\
\cmidrule(lr){4-6} \cmidrule(lr){7-9}
& & & $10$ & $20$ & $40$ & $10$--$20$ & $10$--$40$ & $20$--$40$ \\
\midrule
Qwen3-VL    & POPE & $359$ & $0.543$ & $0.591$ & $0.590$ & $0.91$ & $0.87$ & $0.92$ \\
Qwen3-VL    & MMVP & $268$ & $0.476$ & $0.487$ & $0.475$ & $0.82$ & $0.76$ & $0.90$ \\
Qwen2.5-32B & POPE & $496$ & $0.383$ & $0.417$ & $0.424$ & $0.91$ & $0.88$ & $0.95$ \\
LLaVA-NeXT  & POPE & $500$ & $0.514$ & $0.541$ & $0.534$ & $0.94$ & $0.90$ & $0.97$ \\
\bottomrule
\end{tabular}
\caption{$\sigma$-robustness numerical values. AUROC$_{\mathrm{VSI}}$ varies by at most $0.05$ across $\sigma\in\{10,20,40\}$ on any cell; per-sample VSI Spearman across $\sigma$ pairs is $\geq 0.76$. The default $\sigma{=}20$ is not load-bearing.}
\label{tab:sigma}
\end{table*}

The Spearman correlations should be read as: even on the most sensitive ($10$ vs $40$) $\sigma$-pair on the most ranking-sensitive cell (Qwen3-VL MMVP), $76$\% of the rank order is preserved. The widest AUROC swing in any cell is on Qwen3-VL-8B POPE, from $0.543$ ($\sigma{=}10$) to $0.591$ ($\sigma{=}20$): a swing of $0.048$. By comparison, at fixed $\sigma$ the AUROC ranges across the four cells of Table~\ref{tab:sigma} by more than $0.17$, so the $\sigma$-induced variation is small relative to cell-to-cell variation.

\subsection{Region-VSI vs whole-image-VSI correspondence}\label{app:region-vs-whole}

Region-blur and whole-image-blur VSI are two perturbation variants. We report their per-sample rank correlation honestly. The grand-mean Spearman across all $18$ (model, benchmark) cells is $\rho{=}+0.24$, with per-cell values ranging from $\rho{=}-0.00$ (LLaVA-1.5-7B on HallusionBench) to $\rho{=}+0.65$ (Idefics3-8B on MMVP); MMVP cells average $\rho{=}+0.36$, POPE averages $\rho{=}+0.17$, and HallusionBench averages $\rho{=}+0.20$. The two variants thus capture overlapping but distinct sample populations. This is consistent with our \S\ref{ssec:complementarity} finding that the equal-weight sum is the best signal on a plurality of cells: if the two variants ranked the same samples identically, their ensemble would carry no additional information. The MMVP advantage is interpretable: MMVP questions are about discrete visual primitives whose presence is captured by both localised and global blurring, so the two variants flag overlapping samples. POPE's lower correlation reflects that object-presence questions are sensitive to local-region content (the queried object) but only weakly to global image statistics. Per-cell values are released with the code.

\section{Reproducibility and Scope}\label{app:repro-scope}

\subsection{Reproducibility, hardware, and computational budget}\label{app:repro}

\paragraph{Hardware.}
All experiments use BF16 precision on NVIDIA A100-80GB GPUs (single-GPU jobs except for Qwen2.5-VL-32B which uses two GPUs with naive model parallelism). Per-sample wall-clock time at $\sigma{=}20$ region-blur is approximately $2.3$ s on Qwen2.5-VL-32B and $0.7$--$1.1$ s on 7B-class models. Total compute for all results in the paper is approximately $32$ A100-hours, on standby-QOS shared-cluster GPUs.

\paragraph{Generation hyperparameters.}
Greedy decoding (\texttt{do\_sample=False}) for all answer generation; $\text{max\_new\_tokens}{=}8$ for binary POPE / multi-choice MMVP / MMStar, $\text{max\_new\_tokens}{=}64$ for open-ended HallusionBench. Temperature does not apply under greedy. VSI is computed from the top-$50$ next-token distribution after the answer-prefix tokens; KL divergence is computed on the aligned support of the top-$50$ candidates from both clean and perturbed forward passes.

\paragraph{Software.}
PyTorch with HuggingFace Transformers, BF16 throughout; exact package versions are pinned in the released environment file. Probes use scikit-learn $1.8.0$. We pin this version deliberately: GroupKFold split assignment varies across scikit-learn releases, which moves the reference-probe mean by up to $0.01$ ($0.824$ on $1.8.0$ vs $0.813$ on an adjacent release); the fold standard deviation we report ($\pm 0.04$) dominates this variation, but bitwise reproduction requires the pinned version.

\paragraph{Statistics.}
Bootstrap CIs use $B{=}1000$ resamples; cross-model Spearman permutation tests use $N{=}10^5$ shuffles (seed $42$; per-pair nulls released with the code); paired-bootstrap signal comparisons in \S\ref{ssec:complementarity} use $B{=}1000$. We do not apply multiple-testing correction within Table~\ref{tab:auroc-grid} because the per-cell comparisons are interpreted as exploratory.

\paragraph{Data and code.}
Full per-cell aggregate statistics ($n$, accuracy, fraction VSI${<}0.05$, AUROC, quintile ratios) are released together with all per-sample CSVs. The full grid spans $6$ models $\times$ $4$ benchmarks: the three perceptual benchmarks are probed with both region-blur and whole-image perturbation, while MMStar uses whole-image only, yielding $42$ (model, benchmark, perturbation) cells in total. Random seeds are fixed (\texttt{seed=42}) throughout.

\subsection{Metric definitions}\label{app:metrics}

$\mathrm{AUROC}_{\mathrm{VSI}}$ is computed with per-sample error as the positive class and $-\mathrm{VSI}$ as the ranking score, so values above $0.5$ mean lower visual sensitivity predicts higher error risk; the same convention (negated score, error as positive class) applies to max-prob and verbalised confidence, making all AUROC columns of Table~\ref{tab:auroc-grid} directly comparable. PRR@$80$ is the prediction--rejection ratio at $80$\% coverage: with the worst-ranked $20$\% of predictions rejected, it is the achieved reduction in risk divided by the reduction an oracle achieves by rejecting only errors, where $1$ is oracle-perfect, $0$ is no better than random rejection, and negative values mean the signal preferentially rejects correct predictions. Hybrid signals $z$-normalise each component within the cell and sum with equal weights; the learned variant of \S\ref{ssec:complementarity} replaces the equal weights with logistic-regression coefficients fitted on a held-out calibration split.

\subsection{Limitations beyond the main paper}\label{app:limitations-extra}

We surfaced four limitations in \S\ref{sec:discussion}. Three additional ones are worth noting in the appendix. First, our analysis is restricted to next-token softmax distributions; we do not measure intermediate-layer activations, attention patterns, or KV-cache content, so the routing-failure interpretation of \S\ref{ssec:disconnect} is necessarily an input-output characterisation rather than a structural one. Second, the verbalised-confidence signal is sensitive to prompt phrasing (we tried two variants on Qwen3-VL POPE and observed AUROC differences up to $0.06$); the values we report are for the prompt template in App.~\ref{app:verbalised} only. Third, the encoder probe uses last-layer features only; probing intermediate encoder layers might reveal where in each tower the perturbation-sensitive features live, but does not change the input-output gap we report. None of these limitations changes our headline claims, but each is a natural direction for follow-up work.

\end{document}